\documentclass[final]{cvpr}

\usepackage{times}
\usepackage{epsfig}
\usepackage{graphicx}
\usepackage{amsmath}
\usepackage{amssymb}

\usepackage{rotating}

\usepackage{arydshln}
\usepackage{booktabs}
\usepackage{enumitem}
\usepackage{balance}
\usepackage{combelow}
\usepackage{tabularx}
\usepackage{pifont}
\newcommand{\cmark}{\ding{51}}%

\usepackage{graphicx}
\usepackage{wrapfig}
\usepackage[stable]{footmisc}

\usepackage{multirow}   
\usepackage{makecell}
\usepackage{algorithm}
\usepackage[noend]{algpseudocode}
\usepackage{algorithmicx,eqparbox,array}

\usepackage{xcolor,colortbl}

\usepackage{subcaption}
\usepackage{caption}

\usepackage{mwe}
\usepackage{calc}

\usepackage[pagebackref=true,breaklinks=true,colorlinks,bookmarks=false]{hyperref}

\def\cvprPaperID{4198} 
\def\confYear{CVPR 2021}

\begin{document}

\title{
    Factorizing Perception and Policy for Interactive Instruction Following \\
    \vspace{0.3em}
    \small{\url{https://github.com/gistvision/moca}}
    \vspace{-1em}
}

\author{\hspace{-0.5em}
    Kunal Pratap Singh$^{*,2,\S}$\hspace{0.5em}
    Suvaansh Bhambri$^{*,1}$\hspace{0.5em}
    Byeonghwi Kim$^{*,1}$\hspace{0.5em}
    Roozbeh Mottaghi$^{2}$\hspace{0.5em}
    Jonghyun Choi$^{1,\dagger}$\vspace{0.3em}\\
    {$^1$GIST, South Korea\hspace{10em}$^2$Allen Institute for AI}\\
    {\tt\footnotesize kunals@allenai.org, suvaansh2008bhambri@gmail.com, byeonghwikim@gm.gist.ac.kr}\\
    {\tt\footnotesize roozbehm@allenai.org, jhc@gist.ac.kr}
}

\maketitle

\newcommand\blfootnote[1]{%
  \begingroup
  \renewcommand\thefootnote{}\footnote{#1}%
  \addtocounter{footnote}{-1}%
  \endgroup
}
\begin{abstract}
Performing simple household tasks based on language directives is very natural to humans, yet it remains an open challenge for AI agents.
The `interactive instruction following' task attempts to make progress towards building agents that jointly navigate, interact, and reason in the environment at every step.
To address the multifaceted problem, we propose a model that factorizes the task into interactive perception and action policy streams with enhanced components and name it as \textbf{MOCA}, a Modular Object-Centric Approach.
We empirically validate that MOCA outperforms prior arts by significant margins on the ALFRED benchmark with improved generalization.\blfootnote{\hspace{-2.3em}$^*$: equal contribution. $^\S$: work done while with GIST. $^\dagger$: corresponding author.} 

\vspace{-0.5em}
\end{abstract}

\section{Introduction}
The prospect of having a robotic assistant that can carry out daily chores based on language directives is a distant dream that has eluded the research community for decades. 
On recent progress in computer vision, natural language processing and embodiment, several benchmarks have been developed to encourage research on various components of such instruction following agents including navigation~\cite{anderson2018vision,chen2019touchdown,chaplot2017gated,krantz2020navgraph}, object interaction~\cite{zhu2017visual,misra2017mapping}, and interactive reasoning~\cite{embodiedqa,gordon2018iqa} in  visually rich 3D environments~\cite{ai2thor,chang2017matterport3d,habitat19iccv}. 
However, to move towards building realistic assistants, the agent should possess all these abilities.
Taking a step forward, we address the more holistic task of \textit{interactive instruction following}~\cite{gordon2018iqa,zhu2017visual,shridhar2020alfred,misra2017mapping} which requires an agent to navigate through an environment, interact with objects, and complete long-horizon tasks, following natural language instructions with egocentric vision.

\begin{figure}[t]
     \centering
     \includegraphics[width=.9\columnwidth]{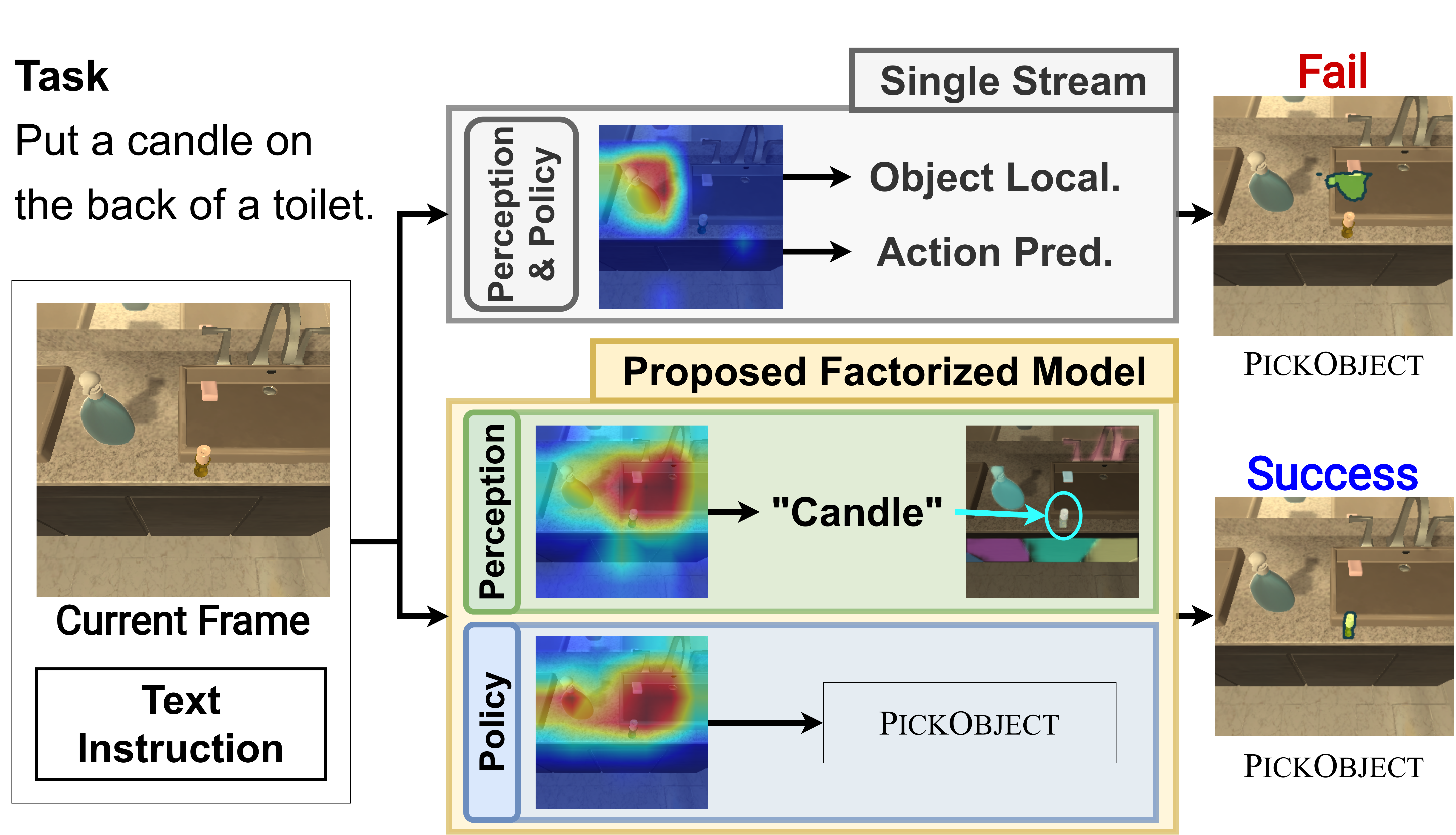}
     \vspace{-0.7em}
     \caption{
         We divide interactive instruction following into perception and policy.
         Each heat-map indicates where a stream focuses on in the given visual observation.
         While a single stream exploits the same features for pixel-level and global understanding and thus fails to interact with the object, our factorized approach handles perception and policy separately and interacts successfully.
     }
     \vspace{-1em}
     \label{fig:teaser}
\end{figure}

To accomplish a goal in the interactive instruction following task, the agent should infer a sequence of actions and object interactions.
While action prediction requires global semantic cues, object localisation needs a pixel-level understanding of the environment, making them semantically different tasks.
In addition, in neuroscience literature~\cite{goodale1992separate}, there is a human visual cortex model that has two pathways; the ventral stream (involved with object perception) and the dorsal stream (involved with action control).
Inspired by them, we present a \textbf{M}odular \textbf{O}bject-\textbf{C}entric \textbf{A}pproach (\textbf{MOCA}) to factorize interactive perception and action policy in separate streams in a unified end-to-end framework for building an interactive instruction following agent.
Specifically, our agent has the action policy module (APM) which is responsible for sequential action prediction and the interactive perception module (IPM) that localises the objects to interact with.

Figure~\ref{fig:teaser} shows that our two-stream model is more beneficial than the single-stream one. 
The heat maps indicate the model's visual attention. 
For the action of `picking up the candle,' the proposed factorized model focuses on a candle in both the streams and results in a successful interaction. 
In contrast, the single-stream model does not attend on the candle, implying the challenge to handle two different predictions in a single stream.

In the IPM, we propose to reason about object classes for better localisation and name it object-centric localisation (OCL). 
We further improve the localising ability in time by using the spatial relationship amongst the objects that are interacted with over consecutive time steps. 
For better grounding of visual features with textual instructions, we propose to use dynamic filters~\cite{Jia2016DynamicFN,landi2019embodied} for its effectiveness of cross-modal embedding.
We also show that these components are more effective when employed in a model that factorizes perception and policy. 

We train our agent using imitation learning, specifically behavior cloning. 
When a trained agent's path is blocked by immovable objects like walls, tables, kitchen counters, \etc. at inference, however, it is likely to fail to escape such obstacles since the ground truth contains only perfect expert trajectories that finish the task without any errors. 
To avoid such errors, we further propose an obstruction evasion mechanism in APM. 
Finally, we adopt data augmentation to address the sample insufficiency of imitation learning. 

We empirically validate our proposed method on the recently proposed ALFRED benchmark~\cite{shridhar2020alfred} and observe that it significantly outperforms prior works in the literature by large margins in all evaluation metrics.

\vspace{0.1em}
We summarize our contributions as follows:
\vspace{-0.7em}
\begin{itemize}
\setlength\itemsep{-0.3em}
   \item We propose to factorize perception and policy for embodied interactive instruction following tasks.
   \item We also present an object-centric localisation and an obstruction evasion mechanism for the task.
   \item We show that this agent outperforms prior arts by large margins in all metrics.
   \item We present qualitative and quantitative analysis to demonstrate our method's effectiveness.

\end{itemize}

\section{Related Work}

\paragraph{Embodied Instruction Following.} Vision and language navigation tasks require an agent to reach a goal by following natural or templated language instructions in a simulated environment through visual observations \cite{anderson2018vision,chaplot2017gated,chen2019touchdown,MacMahon2006WalkTT}.~\cite{anderson2018vision} proposed the Vision-and-Language Navigation (VLN) task on the Room2Room (R2R) benchmark where an agent navigates on a fixed underlying navigation graph based on natural language instructions. Substantial improvements~\cite{wang2018look,fried2018speaker,ma2019regretful,ke2019tactical,Li2019RobustNW,landi2019embodied} have been achieved on this benchmark by various proposals such as progress monitoring~\cite{ma2019selfmonitoring}, augmenting trajectories~\cite{fried2018speaker} and environment dropout~\cite{tan2019learning}. 
Vision and Language Navigation in Continuous Environments (VLN-CE)~\cite{krantz2020navgraph} lifts the assumption of known navigation-graph and perfect agent localisation from R2R~\cite{anderson2018vision}.
Recently,~\cite{ALFWorld20} presented ALFWorld which contains TextWorld~\cite{Ct2018TextWorldAL} based environments corresponding to the ones in~\cite{shridhar2020alfred} that allows agents to learn in an abstract space before transfer to actual embodied environments.

Interactive Instruction Following is a much complex paradigm that combines the navigation aspect of tasks like VLN with the interactive abilities of a manipulation agent~\cite{Batra2020RearrangementAC}. The recently introduced ALFRED~\cite{shridhar2020alfred} benchmark serves as a suitable testbed for this task. It requires an agent to navigate via egocentric visual observations and also interact with objects by producing a pixel-wise mask to complete a task in an embodied environment.
Shridhar \etal~\cite{shridhar2020alfred} proposed a single-stream Seq-to-Seq model with progress monitoring~\cite{ma2019selfmonitoring} for this task. Even though similar models perform well on VLN~\cite{anderson2018vision,ma2019selfmonitoring}, it fails to generalize to unseen environments in interactive instruction following task indicating its difficulty and need of extensive investigation to develop a well-performing agent.
~\cite{saha2021modular} present a planner-based geometry-aware approach for the task. However they split the training data itself to create train, validation and test folds, and do not have any open source code or splits, due to which we omit them in comparison.  
Recently, Nguyen \etal ~\cite{ngyuen_eval_winner} presented an approach wherein they relax the egocentric vision constraint of the task by collecting multiple views per time step, essentially making it similar to panoramic views in VLN~\cite{anderson2018vision}.  They process these visual features via hierarchical attention with the instructions.
Here, we propose to factorize the task into perception and policy to effectively learn an agent for this task. 
Note that we do not relax any constraint set of the original ALFRED benchmark and still outperform prior arts~\cite{shridhar2020alfred,ngyuen_eval_winner}.

\vspace{-1em}\paragraph{Two-stream Architectures.}
\cite{Simonyan2014TwoStreamCN,Cao2019CrossEnhancementTT,feichtenhofer2016convolutional,tesfaldet2018two} have shown the success of multi-stream architectures for capturing different features from given inputs. 
Inspired by these, we also propose a two-stream architecture. Contrary to these works, we do not combine the streams to produce a single output but perform two semantically different tasks i.e. Interactive Perception and Action Policy.
Recently~\cite{jain2021gridtopix,chen2020learning} decouple learning embodied tasks into two parts. Firstly, a perfect perception policy is trained using gridworlds~\cite{jain2021gridtopix} or giving direct access to the environment's state~\cite{chen2020learning}. This is followed by, training the agent on a visually realistic environment to see by imitating the perfect perception policy.



\begin{figure*}[t!]
    \centering
    \includegraphics[width=\linewidth]{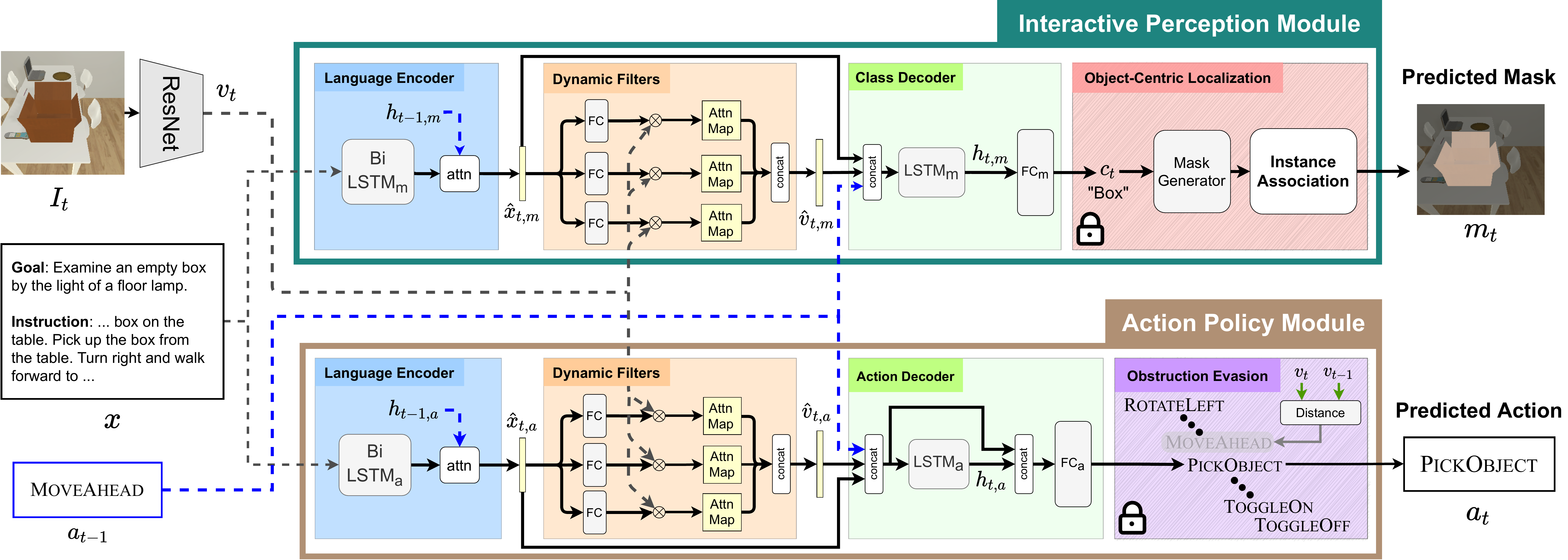}
    \caption{
        \textbf{Model Architecture.}
        The input frame at the time step, $t$, and language instructions are denoted by $I_t$ and $x$.
        {\color{blue}Blue dashed lines} denote the path of the action at the previous time step.
        Subscripts $m$ and $a$ denote that a component belongs to IPM or APM, respectively.
        ResNet-18 encodes $I_t$, denoted by $v_t$.
        Dynamic filters convolve over visual features, $v_t$, to give attended visual features, $\hat{v}_{t,m}$ and $\hat{v}_{t,a}$.
        $h_{t,m}$ and $h_{t,a}$ denote the hidden states of the class and action decoder.
        The target class, $c_t$, and the action, $a_t$, are predicted based on attended visual and language features with the previous action.
        The `lock' symbols in the components indicate their use at inference only.
    }
    \vspace{-1em}
    \label{fig_overview}
\end{figure*}

\vspace{-1em}\paragraph{Visual Grounding.}
Previous visual grounding methods leverage a pre-trained segmentation model~\cite{he2017mask,Zitnick2014EdgeBL,hu2016natural,hu2017modeling,wang2016learning,chen2017query} to generate a set of candidate regions and then predict the best candidate proposal corresponding to the language query.
However, these works have been used for localising a single object in one image with a given language description.
We extend this to the embodied domain and localise multiple objects in a continuous stream of visual observations given a set of instructions. We propose to split the object localisation into two stages; object class prediction and mask generation (Sec.~\ref{section:two_stage}) and leverage a pre-trained instance segmentation model~\cite{he2017mask}.
This is in contrast with~\cite{shridhar2020alfred} which upsamples a vision-language-action embedding via deconvolution layers to produce a class-agnostic mask. As we show in subsequent sections, this results in poorly localised masks.

Previous works~\cite{shridhar2020alfred} have used concatenation for combining vision and language embeddings. However, this fails to fully capture the cross-modal correspondence.
\cite{landi2019embodied} produces dynamic convolution filters that are applied to panoramic visual features to produce action outputs for VLN on R2R benchmark. Motivated by their work, we use dynamic filters for grounding language features with egocentric visual features for interactive instruction following.

\section{Approach}

An interactive instruction following agent performs a sequence of navigational steps and object interactions based on egocentric visual observations it receives from the environment. 
These actions and interactions are based on natural language instructions that the agent must follow to accomplish the task.

We approach this by factorizing the model into two streams, \ie interactive perception and action policy, and train the entire architecture in an end-to-end fashion. 
Figure~\ref{fig_overview} presents a detailed overview of MOCA.

\subsection{Factorizing Perception and Policy}
\label{sec:decouple}
Action prediction requires global scene-level understanding of the visual observation to abstract it to a resulting action. 
On the other hand, for object interaction, the agent needs to focus on both scene-level and object-specific features to achieve precise localisation~\cite{Simonyan2014TwoStreamCN,Liu2018SibNetSC,Cao2019CrossEnhancementTT}.

Given the contrasting nature of the two tasks, MOCA has separate streams for action prediction and object localisation. 
The two streams are the Interactive Perception Module (IPM) and Action Policy Module (APM).
Subscripts $a$ and $m$ in following equations indicate whether a component belongs to APM or IPM, respectively.
APM is responsible for sequential action prediction. It takes in the instructions to exploit the detailed action-oriented information.
IPM localises the pixel-wise mask whenever the agent needs to interact with an object in case of manipulation actions. IPM tries to focus more on object-centric information in the instructions for localisation and interaction.
Both IPM and APM receive the egocentric visual observation features at every time step.

\subsection{Interactive Perception Module (IPM)}
\label{sec:perception}
The ability to interact with objects in the environment is key to interactive instruction following, since accomplishing each task requires multiple interactions.
The interactive perception module (IPM) facilitates this by predicting a pixel-wise mask to localise the object to interact with.

First, the language encoder in IPM encodes the language instructions and generates attended language features.
For grounding the visual features to the language features, we use language guided dynamic filters for generating the attended visual features (Sec.~\ref{section:dynamic_filter}).
Then, to temporally align the correct object with their corresponding interaction actions amongst the ones present in the language input, we use previous action embedding along with the visual and language input.
For example, in the statement, \textit{Wash the spatula, put it in the first drawer}, the agent first needs to wash the spatula in the sink, for which we have two object classes, namely \textit{spatula} and \textit{sink} that the agent needs to interact with. But this has to be done in a particular order. If the action is \textsc{PutObject}, the agent needs to predict the sink's mask whereas if it is \textsc{PickObject}, it needs to predict the spatula's mask. As shown in Figure~\ref{fig_overview}, the hidden state $h_{t,m}$ of the class decoder, $\text{LSTM}_m$, is updated with three different inputs concatenated as:
\begin{equation}
    \begin{split}
        h_{t,m} &= \text{LSTM}_m([\hat{v}_{t,m}; \hat{x}_{t,m}; a_{t-1}])
        \label{eq:sep_class}
    \end{split}
\end{equation}
where $[$;$]$ denotes concatenation. $\hat{x}_{t,m}$ and $\hat{v}_{t,m}$ are the attended language and visual features, respectively.
Finally, the class decoder's current hidden state $h_{t,m}$ is then used to predict the mask $m_t$. 
This is done by invoking the \textit{object-centric localisation} (Sec.~\ref{section:two_stage}), which helps the agent to accurately localise the object of interest.

\subsubsection{Language Guided Dynamic Filters}
\label{section:dynamic_filter}
Visual grounding helps the agent to exploit the relationships between language and visual features. 
This reduces the agent’s dependence on any particular modality while encountering unseen scenarios.

It is a common practice to concatenate flattened visual and language features ~\cite{hu2016natural,shridhar2020alfred,hu2017modeling}. However, it might not perfectly capture the relationship between visual and textual embeddings, leading to poor performance of interactive instruction following agents~\cite{shridhar2020alfred}.

Dynamic filters are conditioned on language features making them more adaptive towards varying inputs. This is in contrast with traditional convolutions which have fixed weights after training and fail to adapt to diverse instructions.
Hence, we propose to use dynamic filters for the interactive instruction following task.

Particularly, we use a filter generator network comprising of fully connected layers to generate dynamic filters which attempt to capture various aspects of the language from the attended language features. 
%
Specifically, the filter generator network, $f_{DF}$, takes the language features, $x$, as input and produces $N_{DF}$ dynamic filters. These filters convolve with the visual features, $v_t$, to output multiple joint embeddings, $\hat{v}_t = \text{DF}(v_t, x)$, as:
\begin{equation}
    \begin{split}
        w_i &= f_{{DF}_i}(x), ~~ i \in [1, N_{DF}], \\
        \hat{v}_{i,t} &= v_t * w_i, \\
        \hat{v}_t &= [\hat{v}_{1,t}; \dots; \hat{v}_{N_{DF},t}],
        \label{eq:df}
    \end{split}
\end{equation}
where $N_{DF}$, $*$ and $[$;$]$ denote the number of dynamic filters, convolution and concatenation operation respectively.
%
We empirically investigate the benefit of using language-guided dynamic filters in Sec.~\ref{sec:model_ablate}.

\subsubsection{Object-Centric Localisation}
\label{section:two_stage}
The IPM performs object interaction by predicting a pixel-wise interaction mask of the object of interest. 
We bifurcate the task of mask prediction; \textit{target class prediction} and \textit{instance association}. This bifurcation enables us to leverage the quality of pre-trained instance segmentation models while also ensuring accurate localisation.
We refer to this mechanism as `object-centric localisation (OCL).'
We empirically validate the OCL in Sec.~\ref{sec:model_ablate} and ~\ref{sec:qual_analysis}.

\vspace{-1em}\paragraph{Target Class Prediction.}
\label{section:target_class}
As the first step of OCL, we take an object-centric viewpoint to interaction by explicitly encoding the ability to reason about object categories in our agent.
To achieve this, MOCA first predicts the target object class, $c_t$, that it intends to interact with at the current time step $t$. 
%
Specifically, $\text{FC}_{m}$ takes as input the hidden state, $h_{t,m}$, of the class decoder and outputs the target object class, $c_t$, at time step, $t$, as shown in Equation~\ref{eq:class_pred}.
The predicted class is then used to acquire the set of instance masks corresponding to the predicted class from the mask generator.
\begin{equation}
    \begin{split} 
        c_t &= \operatorname*{argmax}_k\text{ } \text{FC}_{m}(h_{t,m}), ~~~ k \in [1,N_{class}],
        \label{eq:class_pred}
    \end{split}
\end{equation}
where $\text{FC}_{m}(\cdot)$ is
a fully connected layer and $N_{class}$ denotes the number of the classes of a target object. The target object prediction network is trained as a part of the IPM with the cross-entropy loss with ground-truth object classes. 

\vspace{-1em}\paragraph{Instance Association.}
\label{section:inst_assoc}
At inference, given the predicted object class, we now need to choose the correct mask instance of the desired object.
We use a pre-trained mask generator to obtain the instance masks and confidence scores.
A straightforward solution is to pick the highest confidence instance as it gives the best quality mask of that object.
This works well when the agent interacts with the object for the first time.
However, when it interacts with the same object over an interval, it is more important to \emph{remember} the object the agent has interacted with, since its appearance might vary drastically due to multiple interactions. 
Thus, the sole confidence based prediction may result in failed interactions as it lacks memory.

To address all the scenarios, we propose a two-way criterion to select the best instance mask, \ie, `confidence based' and `association based.'
Specifically, the agent predicts the current time step's interaction mask $m_t = m_{\hat{i},c_t}$ with the center coordinate, $d_t^* = d_{\hat{i}, c_t}$, where $\hat{i}$ is obtained as:
\begin{equation}
    \begin{split}
        \hat{i} = \left\{
            	\begin{array}{ll}
                	\underset{i}{\operatorname{argmax}}~~s_{i,c_t},                   & \mbox{if } c_t \neq c_{t-1},\\
            		\underset{i}{\operatorname{argmin}}~~||d_{i,c_t} - d_{t-1}^*||_2, & \mbox{if } c_t = c_{t-1},
            	\end{array}
            \right.
        \label{eq:instance_association}
    \end{split}
\end{equation}
where $c_t$ is the predicted target object class and $d_{i, c_t}$ the center of a mask instance, $m_{i, c_t}$, of the predicted class.

\begin{figure}[t]
    \centering
    \includegraphics[width=.9\columnwidth]{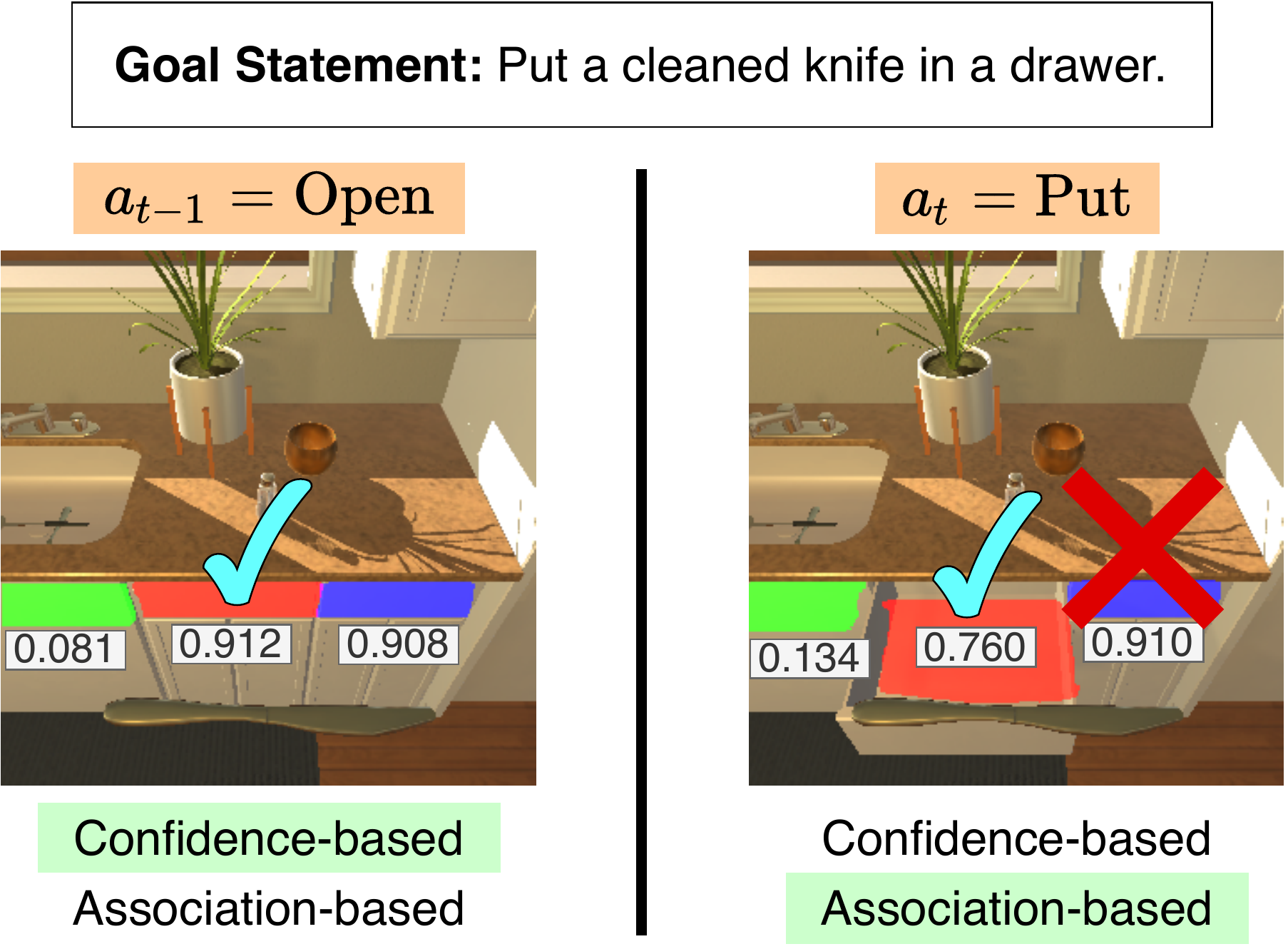}
    \vspace{-0.5em}
    \caption{
        \textbf{Qualitative Illustration of Instance Association (IA).}
        The masks of the drawers are colored with their confidences.~{\color{cyan} \bf $\checkmark$} denotes the object interacted with at that time step.
        {\color{red} \bf $\times$} denotes the object replaced by IA.
        Using the single-fold confidence-based approach could make the agent interact with different drawers since the closed drawer has higher confidence.
        IA helps the agent to interact with the same drawer and place the knife.}
    \vspace{-1.0em}
    \label{fig:closest_instance}
\end{figure}

Figure~\ref{fig:closest_instance} illustrates an example, wherein the agent is trying to open a drawer and put a knife in it, the same drawer is interacted with over multiple time steps.
Table~\ref{tab:ablate_2} in Sec.~\ref{sec:model_ablate} shows ablation study of our instance association scheme.

\subsection{Action Policy Module (APM)}
\label{sec:policy}
The Action Policy Module (APM), depicted by the lower block in Figure~\ref{fig_overview}, is responsible for predicting the action sequence.
It takes visual features and instructions as input.
The attended language features are generated by the language encoder in APM. 
Same as IPM, we employ language guided dynamic filters for generating attended visual features (Sec.~\ref{section:dynamic_filter}). 
Although we use a similar architecture for IPM, the information captured by dynamic filters is different from that of APM due to difference in language encodings used for both.
The action decoder then takes attended visual and language features, along with the previous action embedding to output the action decoder hidden state, $h_{t,a}$. 
Finally, a fully connected layer is used to predict the next action, $a_t$ as follows:
\begin{equation}
    \begin{split}
        u_a = [\hat{v}_{t,a}; \hat{x}_{t,a}; a_{t-1}], ~~~ h_{t,a} = \text{LSTM}_a(u_a) \\
        a_t = \operatorname*{argmax}_k (\text{FC}_{a}([u_a; h_{t,a}]), ~~~ k \in [1, N_{action}]
        \label{eq:sep_action}
    \end{split}
\end{equation}
where $\hat{v}_{t,a}$, $\hat{x}_{t,a}$ and $a_{t-1}$ denote attended visual features, attended language features, and previous action embedding, respectively.
$\text{FC}_a$, takes as input $\hat{v}_{t,a}$, $\hat{x}_{t,a}$, $a_{t-1}$, and $h_{t,a}$ and predicts the next action, $a_t$. Note $N_{action}$ denotes the number of actions. We keep the same action space as~\cite{shridhar2020alfred}. 

The objective function of the APM is the cross entropy with the action taken by expert for the visual observation at each time step as ground-truth.




\vspace{-1em}\paragraph{Obstruction Evasion.}
\label{sec:obstruct_det}
The agent learns to not encounter any obstacles during training based on the expert ground truth actions.
However, during inference, there are various situations when the agent gets stranded around immovable objects.
To address such unanticipated situations, we propose an `obstruction evasion' mechanism in the APM to avoid obstacles at inference time.

While navigating in the environment, at every time step, the agent computes the distance between visual features at the current time step, $v_t$, and the previous time step, $v_{t-1}$ with a tolerance hyper-parameter $\epsilon$ as following: 
\begin{equation}
\label{eq:ob_det}
        d(v_{t-1}, v_t) < \epsilon,
\end{equation}
where $d(v_{t-1}, v_t) = ||v_{t-1} - v_t||^2_2$.
%
When this equation holds, the agent removes the action that causes the obstruction from the action space so that it can escape:
\begin{equation}
    a_t = \mathop{\mathrm{argmax}}\limits_{k} 
    \text{FC}_{a}([u_a; h_{t,a}]), ~~~ k \in [1,N_{action}] - \{k'\}
    \label{eq:OD_2}
\end{equation}
where $k'$ is the index of $a_{t-1}$. $u_a$ and $\text{FC}_a$ are same as Equation~\ref{eq:sep_action}.
We empirically investigate its effect in Sec.~\ref{sec:model_ablate}.

\begin{figure}[t]
    \centering
    \includegraphics[width=.9\columnwidth]{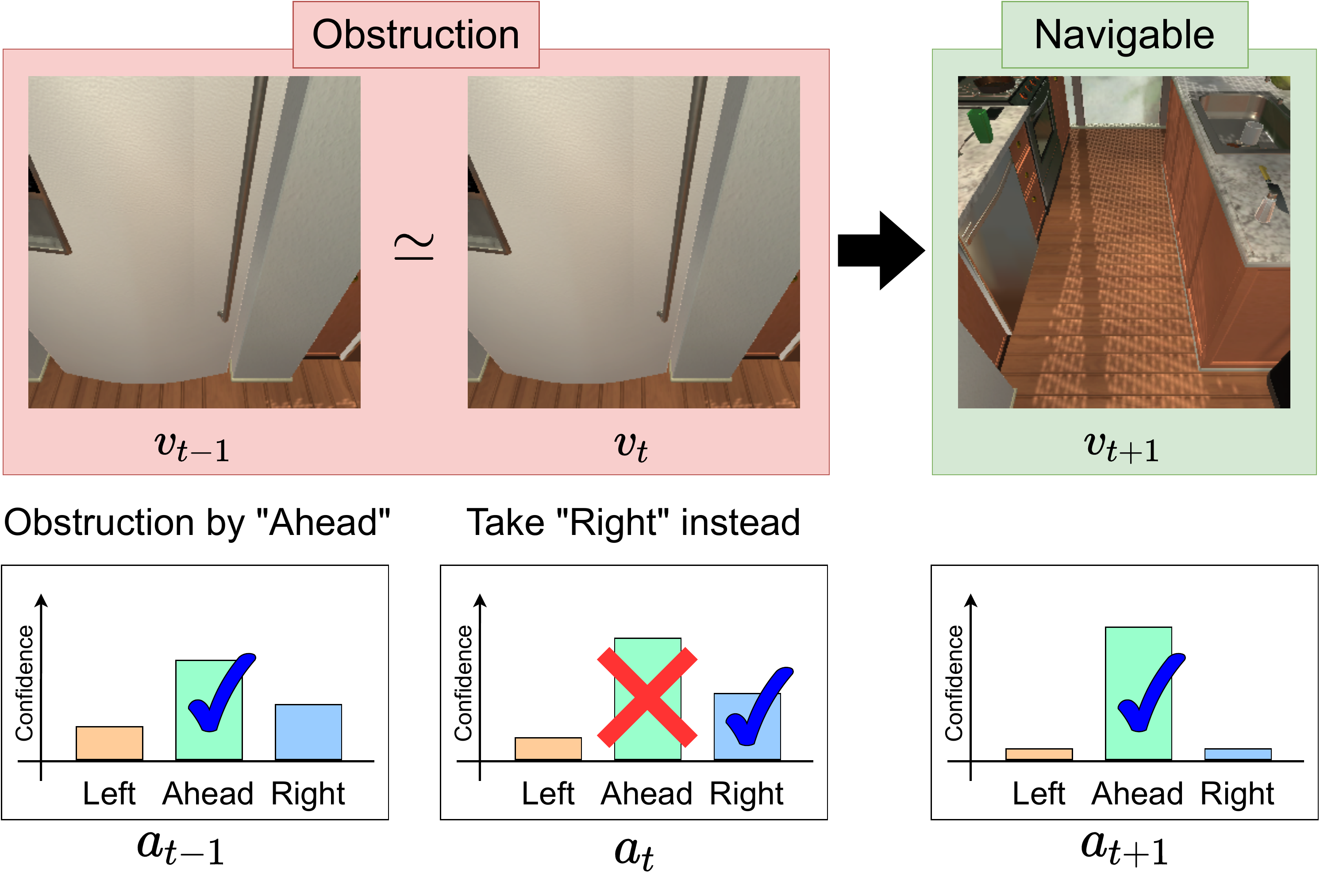}
    \vspace{-1.0em}
    \caption{
        \textbf{Obstruction Evasion}.
        Each plot includes the actions with the top-3 probabilities.
        {\color{blue} \bf $\checkmark$} denotes the action taken at that time step.
        $\textsc{Ahead}$ with {\color{red} \bf $\times$} shows that our agent detects an obstruction at the time step, $t$, by the criteria in Equation~\ref{eq:ob_det}.
        Therefore, our agent predicts the second best action, $\textsc{Right}$, to escape by removing $\textsc{Ahead}$ from the action space.
    }
    \vspace{-1.0em}
    \label{fig:randnav}
\end{figure}

\newcommand{\mcc}[2]{\multicolumn{#1}{c}{#2}}
\newcommand{\mcp}[2]{\multicolumn{#1}{c@{\hspace{30pt}}}{#2}}
\definecolor{Gray}{gray}{0.90}
\newcolumntype{a}{>{\columncolor{Gray}}r}
\newcolumntype{b}{>{\columncolor{Gray}}c}
\newcommand{\B}[1]{\textcolor{blue}{\textbf{#1}}}

\begin{table*}[t!]
    \centering
    \resizebox{1.00\textwidth}{!}{
        \begin{tabular}{@{}laarrcaarr@{}}
            \toprule
            \multirow{3}{*}{Model}
                             & \mcp{4}{\textbf{Validation}} & & \mcc{4}{\textbf{Test}} \\
                             & \mcc{2}{\textit{Seen}}   & \mcc{2}{\textit{Unseen}}  
                             & 
                             & \mcc{2}{\textit{Seen}}   & \mcc{2}{\textit{Unseen}}  \\
                             & \multicolumn{1}{b}{Task} & \multicolumn{1}{b}{Goal-Cond} 
                             & \multicolumn{1}{c}{Task} & \multicolumn{1}{c}{Goal-Cond} 
                             & 
                             & \multicolumn{1}{b}{Task} & \multicolumn{1}{b}{Goal-Cond} 
                             & \multicolumn{1}{c}{Task} & \multicolumn{1}{c}{Goal-Cond} \\
            \cmidrule{1-5} \cmidrule{7-10}
            {Shridhar \etal~\cite{shridhar2020alfred}}        & $3.70$ ($2.10$)    & $10.00$  ($7.00$)    & $0.00$ ($0.00$)   & $6.90$ ($5.10$)  & & $3.98$ ($2.02$)   & $9.42$ ($6.27$)   & $0.39$ ($0.80$) & $7.03$ ($4.26$) \\[1pt]
            {Nguyen \etal~\cite{ngyuen_eval_winner}}          & \multicolumn{1}{b}{N/A} & \multicolumn{1}{b}{N/A} & \multicolumn{1}{c}{N/A} & \multicolumn{1}{c}{N/A}      & & $12.39$ ($8.20$) & $20.68$ ($18.79$)   & $4.45$ ($2.24$) & $12.34$ ($9.44$) \\[1pt]
            {MOCA (Ours)}                                     & $\B{25.85}$ ($\B{18.95}$)    & $\B{34.92}$  ($\B{26.44}$)    & $\B{5.36}$ ($\B{3.19}$)   & $\B{16.18}$ ($\B{10.44}$)        & & $\B{26.81}$ ($\B{19.52}$)   & $\B{33.20}$ ($\B{26.33}$)   & $\B{7.65}$ ($\B{4.21}$) & $\B{15.73}$ ($\B{11.24}$) \\[1pt]
            \cmidrule{1-5} \cmidrule{7-10}
            \multicolumn{10}{l}{\footnotesize \bf {\fontfamily{lmss}\selectfont \hspace{-1em} Input Ablations}}\\
            {~~No Language}          & $2.00$ ($1.59$) & $10.85$ ($5.69$)     & $0.00$ ($0.00$) & $4.11$ ($1.60$) & & $0.59$ ($0.29$) & $6.37$ ($4.24$) & $0.20$ ($0.03$) & $6.82$ ($3.43$) \\ %
            {~~No Vision}            & $0.12$ ($0.05$) & $6.16$ ($5.11$)     & $0.00$ ($0.00$) & $7.26$ ($6.41$) & & $0.07$ ($0.03$) & $4.31$ ($3.34$) & $0.20$ ($0.07$) & $6.92$ ($4.72$) \\
            {~~Goal-Only}            & $3.90$ ($2.59$) & $11.43$ ($8.65$)   & $0.49$ ($0.12$) & $8.40$ ($4.66$) & & $3.59$ ($2.39$) & $10.03$ ($7.47$) & $1.11$ ($0.40$) & $8.70$ ($4.96$) \\
            {~~Instructions-Only}    & $5.98$ ($4.52$) & $14.56$ ($11.16$)    & $0.49$ ($0.27$) & $7.97$ ($5.09$) & & $6.20$ ($3.96$) & $12.44$ ($9.45$) & $0.85$ ($0.36$) & $7.84$ ($4.62$) \\
            \cmidrule{1-5}\cmidrule{7-10}
            {Human}     & \multicolumn{1}{b}{-} & \multicolumn{1}{b}{-} & \multicolumn{1}{c}{-} & \multicolumn{1}{c}{-} & & \multicolumn{1}{b}{-} & \multicolumn{1}{b}{-} & $91.00$ ($85.80$) & $94.50$ ($87.60$) \\
            \bottomrule
        \end{tabular}
    }
    \vspace{-0.5em}
    \caption{\textbf{Task and Goal-Condition Success Rate.}
        For each metric, the corresponding path weighted metrics are given in (parentheses).
        The highest values per fold and metric are shown in \B{blue}.
        `N/A' denotes `not available' as the scores are not reported in the leaderboard. 
    }
    \vspace{-10pt}
    \label{tab:results}
\end{table*}

\section{Experiments}
We present quantitative comparisons and show that we outperform prior works~\cite{shridhar2020alfred,ngyuen_eval_winner}
with large margins. 
We also perform extensive ablation studies and additional analyses over the empirical significance of various components of MOCA and discuss qualitative examples to highlight the importance of our design choices.

\vspace{-1em}\paragraph{Dataset.}
For training and evaluating on the interactive instruction following task, we use the recently proposed ALFRED benchmark that runs in AI2-THOR~\cite{ai2thor}. The  scenes in ALFRED are divided into `train', `validation' and `test' sets. 
To evaluate the generalization ability, the validation and test scenes are split into two sections; \emph{seen} and \emph{unseen} folds.
Scenes in the seen folds of validation and test data are subsets of those in the train fold. Scenes in the unseen validation and test folds are distinct from the train fold and from each other.
The dataset provides both high-level goal statement and low-level step-by-step instructions.
We provide the detailed description of the ALFRED benchmark and our implementation details in the supplementary.

\vspace{-1em}\paragraph{Evaluation Metrics.}
We follow the evaluation metrics proposed in~\cite{shridhar2020alfred}, \ie, Success Rate, denoted by \textit{Task}, and Goal Condition Success Rate, denoted by \textit{Goal-Cond}.
Additionally, to measure the efficiency of an agent, the above metrics are penalized by the length of the path to compute a path-length-weighted (PLW) score for each metric~\cite{Anderson2018OnEO}.
For more details on evaluation metrics, kindly refer ~\cite{shridhar2020alfred}.



\subsection{Quantitative Analysis}
\label{sec:quant_analysis}
We first conduct quantitative analysis of the performance on task success rate (Task) and goal-condition success rate (Goal-Cond) and summarise the results in Table~\ref{tab:results} with previous methods.
As shown in the figure, MOCA shows significant improvement over the prior arts~\cite{shridhar2020alfred,ngyuen_eval_winner} on all metrics. The higher success rate in the unseen scenes indicates its ability to generalize in novel environments. We achieve an improvement of 14.42\% and 3.20\% in Seen and Unseen Task SR over Nguyen \etal~\cite{ngyuen_eval_winner} that won ALFRED challenge in ECCV 2020.
Note that Nguyen \etal~\cite{ngyuen_eval_winner} is a challenge entry, they neither report validation set results nor have a code release, hence the comparison was omitted.

MOCA outperforms them in both \emph{Seen} and \emph{Unseen} `Goal-Condition' metrics and gives an improvement of 12.52\% and 3.39\%, respectively. 
The superior performance of MOCA on both overall task success rate and goal condition indicates its understanding of short-term sub-tasks, as well as long-horizon full tasks. \cite{shridhar2020alfred} lacks long term task completion ability as indicated by its poor performance on Task Success Rate. As indicated in the parenthesis in Table~\ref{tab:results}, MOCA provides better Path Length Weighted results for all the metrics which shows the efficiency of our agent.~
We would also like to acknowledge that in the ALFRED public leaderboard\footnote{https://leaderboard.allenai.org/alfred/submissions/public}, the highest entry is at 9.42 unseen success rate, but it is only an anonymous leaderboard entry with no manuscript or code, thereby we omit it in comparison. We present sub-goal and task type ablations in the supplementary.

\subsection{Ablation Study}
\label{sec:ablation_study}

\paragraph{Input Ablations.}
\label{sec:inp_ablations}
We ablate the inputs to our model in Table~\ref{tab:results} to investigate the vision and language bias of MOCA. When the agent is only given visual inputs (\textit{No Language}) \ie zeroing out language input, we observe that it's able to perform some tasks in the seen environments by memorising familiar visual sequences, but fails to generalize in the unseen environment. 

\textit{No Vision} setting is able to finish some goal conditions by following navigation instructions, but the lack of visual input handicaps the interaction ability of the agent, hence it drastically fails in both seen and unseen folds. 

\textit{Goal-Only} setting highlights the ability of MOCA to utilise the goal-statement better as compared to Shridhar \etal~\cite{shridhar2020alfred}. Since our Action Policy Module (APM) does not utilise the goal-statement as it lacks action-specific information, the action prediction ability of this setting is equivalent to the \textit{No-Language} setting. However, since the goal-statement is used in the Interactive Perception Module (IPM), it allows the agent to perform accurate object interactions and hence achieves much better performance than \textit{No-Language}. This result is a direct benefit of the perception and policy factorization discussed in Sec.~\ref{sec:decouple}.

\textit{Instruction-Only} ablation in Table~\ref{tab:results} indicates the performance when the agent does not receive the goal-statement. The instructions drastically improve the action prediction ability over the \textit{Goal-Only} setting as the APM can now leverage the detailed action information. However, the IPM is deprived of its language input which depletes the target-class prediction ability (Sec.~\ref{section:target_class}) of object-centric localisation. This results in many failed interactions and thus it performs worse than our full model and \textit{Goal-Only} setting.


It is also worth noting that for input ablations, the agent is deprived of the dynamic filters for either APM or IPM, or both, due to which it fails to perform well on unseen environments in all input ablation settings.

\vspace{-1em}\paragraph{Stream Input Ablations.} 
\label{sec:branch_input}
As mentioned before, we use the goal statement as the input to IPM and instructions for the APM for our experiments.
However, we perform an empirical study to show that our framework is not sensitive to this particular choice and can generalize beyond it.
We investigate the language inputs with different goal and instruction combinations in Table~\ref{tab:ablation_lang}.

We replace the input to APM and/or IPM by a concatenation of goal and instructions similar to~\cite{shridhar2020alfred} and report the task success rate on the resulting combinations.
As shown in Table~\ref{tab:ablation_lang}, we do not observe any performance degradation, which indicates that our approach is not sensitive to the choice of language inputs.
Note that it is possible to optimize the choice of language inputs for minor performance gains, but we keep the current combination for the ease of analysis. Moreover, our goal is to contribute a general framework for interactive instruction following task which is agnostic to language instruction type, that can generalize beyond ALFRED~\cite{shridhar2020alfred}.

\begin{table}[t!]
    \centering
    \resizebox{0.48\textwidth}{!}{
        \begin{tabular}{@{}ccbbcccc@{}}
            \toprule
            \multicolumn{2}{c}{\bf Input} & \multicolumn{2}{c}{\bf Val-Seen} & \multicolumn{2}{c}{\bf Val-Unseen} \\
            \cmidrule(lr){1-2} \cmidrule(lr){3-4} \cmidrule(lr){5-6}
            IPM & APM & \multirow{1}{*}{Task}   & \multirow{1}{*}{Goal-Cond.} &  \multirow{1}{*}{Task} &  \multirow{1}{*}{Goal-Cond.} \\
            \cmidrule(lr){1-2} \cmidrule(lr){3-4} \cmidrule(lr){5-6}
            G & I                                               & $25.85$ ($18.95$) & $34.92$ ($26.44$) & $5.36$ ($3.19$)   & $16.18$ ($10.44$)  \\
            \cmidrule(lr){1-2} \cmidrule(lr){3-4} \cmidrule(lr){5-6}
            G,I & I                                 & $29.76$ ($22.33$) & $39.40$ ($30.58$) & $5.97$ ($3.52$) & $18.25$ ($11.78$) \\ 
            G & G,I                              & $28.05$ ($20.96$) & $35.89$ ($28.24$) & $5.36$ ($3.21$) & $17.26$ ($10.56$) \\ 
            G,I & G,I                                 & $26.34$ ($18.20$) & $34.28$ ($25.68$) & $5.36$ ($2.72$)   & $16.23$ ($9.28$)  \\ 
            \bottomrule
        \end{tabular}
    }
    \vspace{-0.5em}
    \caption{
        \textbf{Stream Input Ablations for Interactive Perception Module (IPM) and Action Policy Module (APM).}
        For each metric, we report the corresponding path weighted scores in parentheses.
        Each ``G'' and ``I'' denotes a goal statement and step-by-step instructions and ``G,I'' the concatenation of them.
    }
    \vspace{-0.5em}
    \label{tab:ablation_lang}
\end{table}

\begin{table}[t!]
    \centering
    \resizebox{0.48\textwidth}{!}{
        \begin{small}
            \begin{tabular}{@{}cccccbc@{}}
                \toprule
                \multirow{1}{*}{~~~\#} & FPP & OCL & DF & DA & \multirow{1}{*}{\cellcolor[HTML]{FFFFFF} \textbf{Val-Seen} Task} &  \multirow{1}{*}{\textbf{Val-Unseen} Task} \\
                \cmidrule(lr){1-1} \cmidrule(lr){2-5} \cmidrule(lr){6-6} \cmidrule(lr){7-7}
                \multicolumn{1}{c}{($a$)} & \cmark & \cmark & \cmark & \cmark                                               & $25.85$ ($18.95$) & $5.36$ ($3.19$) \\%
                \multicolumn{1}{c}{($b$)} & \multicolumn{1}{c}{} & \cmark & \cmark & \cmark                                 & $22.32$ ($16.17$) & $4.51$ ($2.59$) \\%
                \cmidrule(lr){1-1} \cmidrule(lr){2-5} \cmidrule(lr){6-6} \cmidrule(lr){7-7}
                \multicolumn{1}{c}{($c$)} & \cmark & \cmark & \cmark & \multicolumn{1}{c}{}                                 & $15.85$ ($10.02$) & $2.92$ ($1.35$) \\%
                \multicolumn{1}{c}{($d$)} & \multicolumn{1}{c}{} & \cmark & \cmark & \multicolumn{1}{c}{}                   & $~~12.56$ ($7.05$) & $2.68$ ($1.32$) \\%
                \multicolumn{1}{c}{($e$)} & \cmark & \cmark & \multicolumn{1}{c}{} & \multicolumn{1}{c}{}                   & $~~14.63$ ($9.80$) & $2.19$ ($1.23$) \\%
                \multicolumn{1}{c}{($f$)} & \multicolumn{1}{c}{} & \cmark & \multicolumn{1}{c}{} & \multicolumn{1}{c}{}     & $~~11.71$ ($5.42$) & $1.83$ ($0.82$) \\
                \multicolumn{1}{c}{($g$)} & \cmark & \multicolumn{1}{c}{} & \cmark & \multicolumn{1}{c}{}                   & $~~~~3.90$ ($2.40$)  & $0.50$ ($0.30$) \\%
                \multicolumn{1}{c}{($h$)} & \multicolumn{1}{c}{} & \multicolumn{1}{c}{} & \cmark & \multicolumn{1}{c}{}     & $~~~~3.30$ ($1.70$)  & $0.40$ ($0.20$) \\%
                \bottomrule
            \end{tabular}
        \end{small}
    }
    \vspace{-0.5em}
    \caption{
        \textbf{Ablation Study for Each Component of the Proposed Model.}
        FPP denotes factorizing perception and policy.
        OCL denotes object-centric Localisation.
        DF denotes language-guided dynamic filters.
        DA denotes data augmentation.
        For each metric, we report task success rates with corresponding path weighted scores in parentheses.
        The absence of checkmark denotes that the corresponding component is removed.
    }
    \vspace{-1em}
    \label{tab:ablation}
\end{table}

\vspace{-1em}\paragraph{Model Ablations.}
\label{sec:model_ablate}

\begin{table}[t!]
    \centering
    \resizebox{0.48\textwidth}{!}{
        \begin{tabular}{@{}laarrcaarr@{}}
            \toprule
            \multirow{2}{*}{Model}
                             & \mcc{2}{\textbf{Val-Seen}}   & \mcc{2}{\textbf{Val-Unseen}}
                             \\
                             & \multicolumn{1}{b}{Task} & \multicolumn{1}{b}{Goal-Cond} 
                             & \multicolumn{1}{c}{Task} & \multicolumn{1}{c}{Goal-Cond} 
                             \\
            \midrule
            {MOCA}   & $25.85$ ($18.95$)   & $34.92$  ($26.44$)     & $5.36$ ($3.19$)   & $16.18$ ($10.44$) \\[1pt]
            \midrule
            {-- w/o I.A.} & $23.66$ ($17.47$)   & $32.48$ ($25.18$)     & $5.12$ ($3.04$)   & $15.85$ ($10.32$) \\
            {-- w/o O.E.} & $20.00$ ($15.08$)  & $28.26$ ($22.67$)     & $3.53$ ($2.38$)   & $14.25$ ($10.53$) \\
            \bottomrule
        \end{tabular}
    }
    \caption{
        \textbf{Ablation for Instance Association and Obstruction Evasion.}
        Both the components are ablated on the validation set.
    }
    \vspace{-1em}
    \label{tab:ablate_2}
\end{table}

\begin{figure*}[t!]
    \centering
    \begin{subfigure}{.48\linewidth}
        \includegraphics[width=\columnwidth]{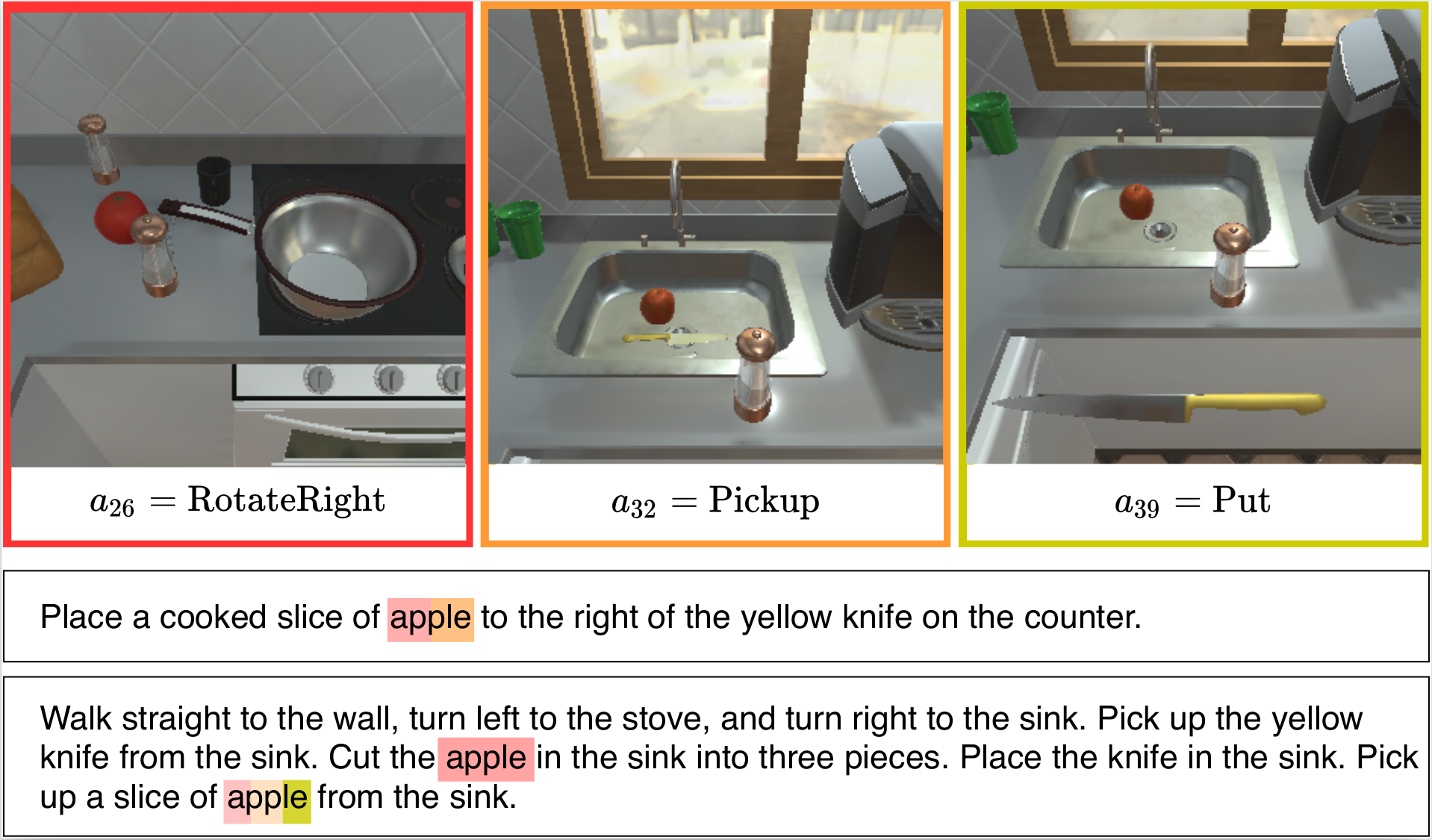}
        \caption{MOCA w/o factorizing perception and policy}
        \label{fig:langAttn_dec_coup}
    \end{subfigure}
    \begin{subfigure}{.48\linewidth}
        \includegraphics[width=\columnwidth]{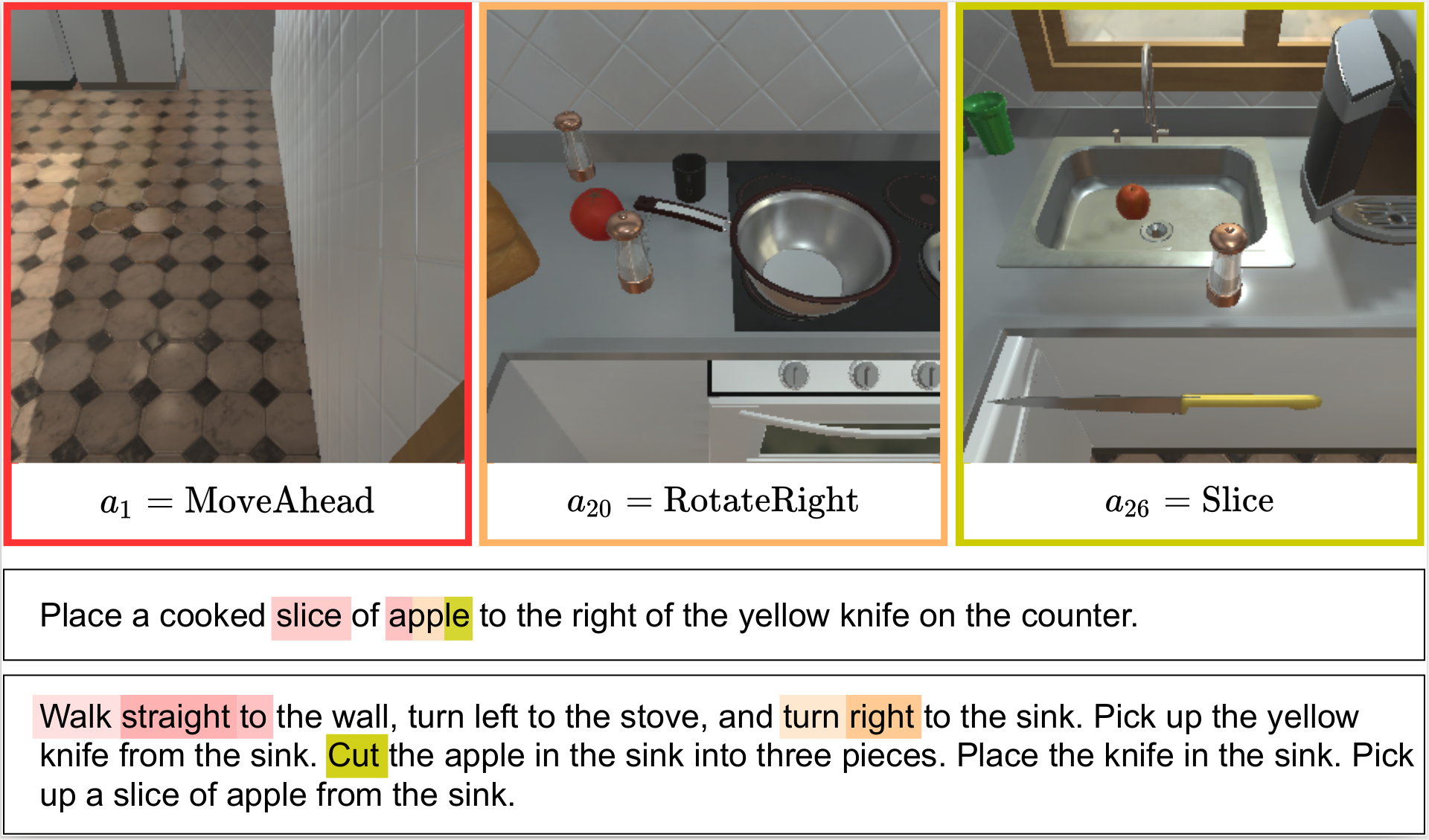}
        \caption{MOCA}
        \label{fig:langAttn_dec_decoup}
    \end{subfigure}
    \caption{
        \textbf{Language Attention for Single-Stream and Two-Stream Models.}
        The colors of frame borders and words denote that the agent at the particular frame focuses on the same-colored words.
        $a_t$ denotes the action taken at the time step, $t$. (a) Without factorization, the language attention keeps focusing on \textit{apple} irrespective of the action taken. (b) With factorization, the language attention focuses on the words that correspond to the action taken at that time step.
    }
    \vspace{-1em}
    \label{fig:langAttn_dec}
\end{figure*}

To investigate the significance of each component with empirical studies, we perform a series of ablations on MOCA and summarize the results in Table~\ref{tab:ablation}.
We only present the task success rate due to space constraints. We present the full table in supplementary.
\# $(a)$ represents our full model.
We begin by showing that factorization is important for models both with (\# $(a)$ vs.\ $(b)$) and without (\# $(c)$ vs.\ $(d)$) data augmentation. 
For this ablation, we take the concatenation of goal and instructions as the language input and perform action and mask prediction from a single stream similar to~\cite{shridhar2020alfred} while keeping other modules the same.
Note that the presence of (\checkmark) in 'FPP' column indicates whether the model is two-stream(\checkmark) or single-stream (no~\checkmark).
We also find that data augmentation is important (\# $(a)$ vs.\ $(c)$) in training a better and more generalizable agent for the task.

Next, we ablate over the language guided dynamic filters (Sec.~\ref{section:dynamic_filter}). 
Removing them leads to a decrease in both seen and unseen metrics (\# $(c)$ vs.\ $(e)$).
This drop can be attributed to the lack of cross-modal correspondence between visual and language inputs.
We also show that dynamic filters are less effective (\# $(g)$ vs.\ $(h)$) without factorization. This solidifies our understanding that a two-stream architecture is better-suited for interactive instruction following.


Finally, we ablate over object-centric localisation (OCL) (Sec.~\ref{section:two_stage}).
We observe that the performance drastically drops (\# $(c)$ vs.\ $(g)$) on both seen and unseen folds due to poor localisation, highlighting the effectiveness of our object-centric design. Note that the large drop also indicates the importance of object localisation, and hence interaction in the task.
Additionally, we also show that OCL is more effective with factorization (\# $(e)$ vs.\ $(f)$), further highlighting the importance of factorization for our agent's superior performance.
For this ablation, to remove OCL, we directly upsample the joint vision-language-action embedding using deconvolution layers to predict the mask similar to~\cite{shridhar2020alfred}.

Table~\ref{tab:ablate_2} ablates obstruction evasion from Sec.~\ref{sec:obstruct_det}. The performance drop indicates that it helps the agent to avoid obstacles effectively. 
We also ablate over the Instance-Association (IA) presented in Sec.~\ref{section:inst_assoc}. For this setting, instead of picking the mask instance for the predicted target class using IAT, we pick a random instance of that class. This setting achieves almost half the performance of MOCA which implies that merely predicting the right object class is not enough, the correct instance must be selected.

\subsection{Qualitative Analysis}
\label{sec:qual_analysis}

\paragraph{Factorizing Perception and Policy.}
We present a qualitative example of the benefit of factorizing perception and policy. 
In Figure~\ref{fig:langAttn_dec_coup}, for the single stream model \ie without factorization, the language attention focuses on the objects mentioned in the goal statement, such as \textit{apple} on all three shown time steps, even though it is not relevant to the current action ignoring all other action-specific information in the instructions. 

However, when perception and policy are factorized and we use a two-stream model, it can effectively encode the representations for both interactive perception and policy.
Therefore, the attention mechanism focuses on the correct words for both navigation and interactive actions. 
Figure~\ref{fig:langAttn_dec_decoup} qualitatively illustrates this result. 
For example, at $t = 20$ MOCA attends over \textit{turn right}  when it predicts \textsc{RotateRight}. At $t = 26$ when our agent intends to slice the apple, it attends over \textit{Cut}.
Note that the only difference between the models is factorization of perception and policy. 

\begin{figure}[t!]
    \centering
    \begin{subfigure}{\linewidth}
       \centering
        \begin{subfigure}{.42\linewidth}
             \centering
             \includegraphics[width=\linewidth]{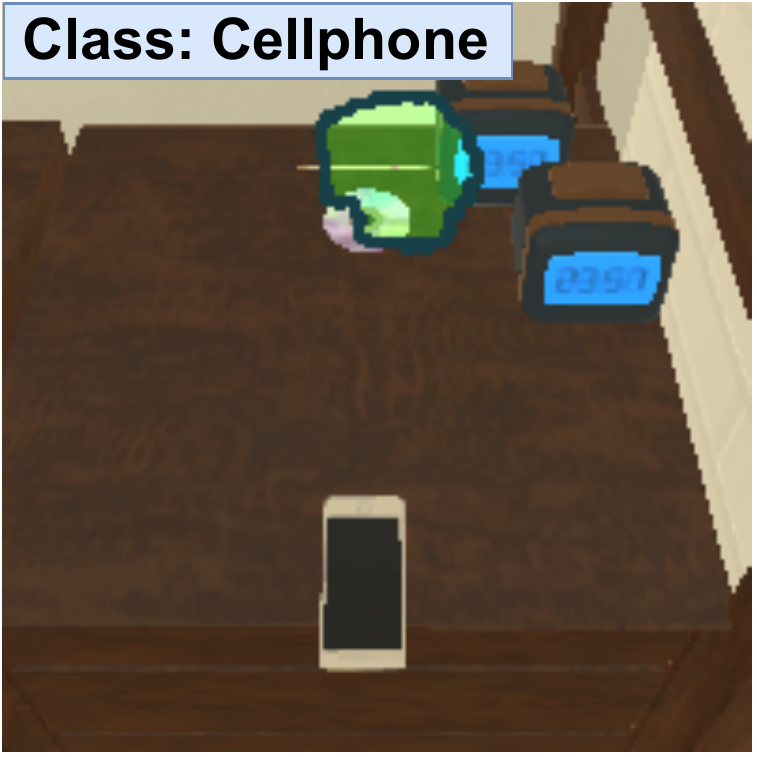}
        \end{subfigure}%
        \hspace{1em}%
        \hspace{0.7em}%
        \begin{subfigure}{.42\linewidth}
             \centering
             \includegraphics[width=\linewidth]{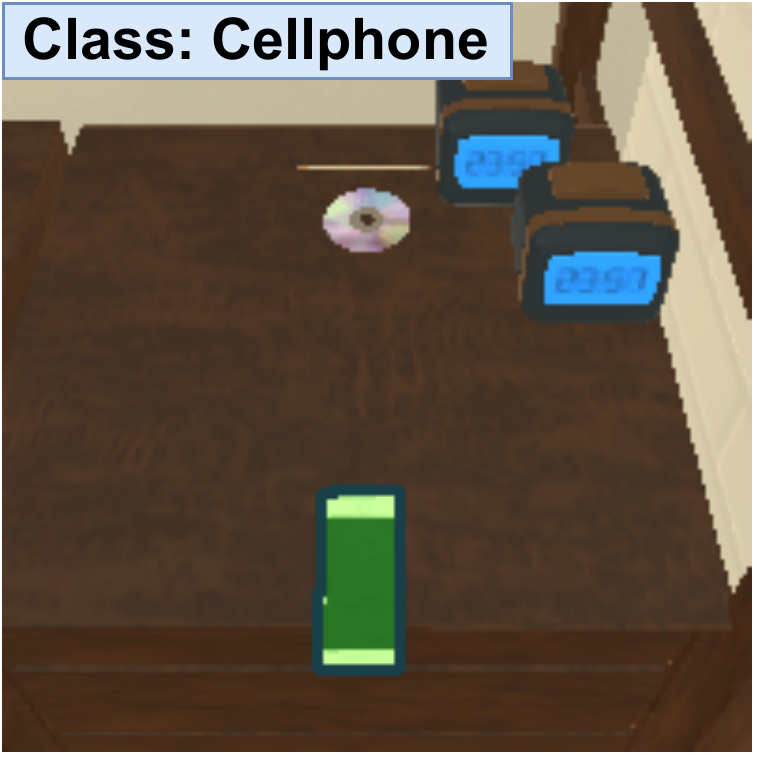}
        \end{subfigure}%
    \end{subfigure}\\%
    \vspace{.5em}
    \begin{subfigure}{\linewidth}
        \centering
        \begin{subfigure}{.42\linewidth}
            \centering
            \caption{MOCA w/o OCL}
            \label{fig_qual_res_1}
        \end{subfigure}%
        \hspace{1.5em}%
        \begin{subfigure}{.42\linewidth}
            \centering
            \caption{MOCA}
            \label{fig_qual_res_2}
        \end{subfigure}%
    \end{subfigure}%
    \vspace{-1.1em}
    \caption{
        \textbf{Qualitative Comparison of Object Localisation.}
        Green regions denote the masks predicted by the models.
        The ground-truth object class the agent needs to interact with is shown on the top-left corner. OCL denotes object-centric localisation.
    }
    \vspace{-1.5em}
    \label{fig_qual_res}
\end{figure}

\vspace{-1em}\paragraph{Object-Centric Localisation.}
We also conduct qualitative analysis of the object localisation ability (Sec.~\ref{section:two_stage}) of MOCA. 
\textit{Object-Centric Localisation} (OCL) allows our method to reason about object classes (Sec.~\ref{section:target_class}) which ensures interaction with the correct object. 
This is in contrast with~\cite{shridhar2020alfred} that upsamples a linear embedding via a deconvolution network and predicts class-agnostic masks, thereby not preserving any information about object category.
In Figure~\ref{fig_qual_res_1}, for Ours w/o OCL setting, we replace OCL by deconvolution layers similar to~\cite{shridhar2020alfred}. Since it lacks the ability to reason about object class, it predicts inaccurate masks even though the objects are fully observable. 

In contrast, in Figure~\ref{fig_qual_res_2}, our full method successfully predicts what objects it intends to interact with (\ie, the cellphone). 
Identifying the correct objects enables it to predict an accurately localised mask with the mask generator's help.
We present further qualitative example videos of MOCA's task completion ability in the supplementary.

\section{Conclusion}

We explore the problem of interactive instruction following. 
To address this compositional task,  we propose a model that factorizes the task into interactive perception and action policy. We also propose improved components for object localisation and obstacle avoidance.
Our method provides a framework that can be adopted by future works on ALFRED and beyond. 
Our approach outperforms all prior arts by significant margins with superior generalization. We present extensive analysis and insights that can benefit the general paradigm of instruction following.

{
\small
\noindent
\textbf{Acknowledgement.} This work was partly supported by the National Research Foundation of Korea (NRF) grant funded by the Korea government (MSIT) (No.2019R1C1C1009283) and Institute of Information \& communications Technology Planning \& Evaluation (IITP) grant funded by the Korea government (MSIT) (No.2019-0-01842, Artificial Intelligence Graduate School Program (GIST)) and (No.2019-0-01351, Development of Ultra Low-Power Mobile Deep Learning Semiconductor With Compression/Decompression of Activation/Kernel Data).\par
}

{\small
\bibliographystyle{ieee_fullname}
\bibliography{cvpr}

\begin{thebibliography}{10}\itemsep=-1pt

\bibitem{Anderson2018OnEO}
Peter Anderson, Angel~X. Chang, Devendra~Singh Chaplot, Alexey Dosovitskiy,
  Saurabh Gupta, Vladlen Koltun, Jana Kosecka, Jitendra Malik, Roozbeh
  Mottaghi, Manolis Savva, and Amir~R. Zamir.
\newblock On evaluation of embodied navigation agents.
\newblock arXiv:1807.06757, 2018.

\bibitem{anderson2018vision}
Peter Anderson, Qi Wu, Damien Teney, Jake Bruce, Mark Johnson, Niko
  S{\"u}nderhauf, Ian Reid, Stephen Gould, and Anton van~den Hengel.
\newblock Vision-and-language navigation: Interpreting visually-grounded
  navigation instructions in real environments.
\newblock In {\em CVPR}, 2018.

\bibitem{Batra2020RearrangementAC}
Dhruv Batra, Angel~X. Chang, Sonia Chernova, Andrew~J. Davison, Jia Deng,
  Vladlen Koltun, Sergey Levine, Jitendra Malik, Igor Mordatch, Roozbeh
  Mottaghi, Manolis Savva, and Hao Su.
\newblock Rearrangement: A challenge for embodied ai.
\newblock arXiv:2011.01975, 2020.

\bibitem{Cao2019CrossEnhancementTT}
Dong Cao and Lisha Xu.
\newblock Cross-enhancement transform two-stream 3d convnets for pedestrian
  action recognition of autonomous vehicles.
\newblock arXiv:1908.08916, 2019.

\bibitem{chang2017matterport3d}
Angel Chang, Angela Dai, Thomas Funkhouser, Maciej Halber, Matthias Niessner,
  Manolis Savva, Shuran Song, Andy Zeng, and Yinda Zhang.
\newblock Matterport3d: Learning from rgb-d data in indoor environments.
\newblock arXiv:1709.06158, 2017.

\bibitem{chaplot2017gated}
Devendra~Singh Chaplot, Kanthashree~Mysore Sathyendra, Rama~Kumar Pasumarthi,
  Dheeraj Rajagopal, and Ruslan Salakhutdinov.
\newblock Gated-attention architectures for task-oriented language grounding.
\newblock In {\em AAAI}, 2017.

\bibitem{chen2020learning}
Dian Chen, Brady Zhou, Vladlen Koltun, and Philipp Kr{\"a}henb{\"u}hl.
\newblock Learning by cheating.
\newblock In {\em CoRL}, 2020.

\bibitem{chen2019touchdown}
Howard Chen, Alane Suhr, Dipendra Misra, Noah Snavely, and Yoav Artzi.
\newblock Touchdown: Natural language navigation and spatial reasoning in
  visual street environments.
\newblock In {\em CVPR}, 2019.

\bibitem{chen2017query}
Kan Chen, Rama Kovvuri, and Ram Nevatia.
\newblock Query-guided regression network with context policy for phrase
  grounding.
\newblock In {\em ICCV}, 2017.

\bibitem{Ct2018TextWorldAL}
Marc-Alexandre C{\^o}t{\'e}, {\'A}kos K{\'a}d{\'a}r, Xingdi Yuan, Ben Kybartas,
  Tavian Barnes, Emery Fine, James Moore, Matthew~J. Hausknecht, Layla~El Asri,
  Mahmoud Adada, Wendy Tay, and Adam Trischler.
\newblock Textworld: A learning environment for text-based games.
\newblock In {\em CGW@IJCAI}, 2018.

\bibitem{embodiedqa}
Abhishek Das, Samyak Datta, Georgia Gkioxari, Stefan Lee, Devi Parikh, and
  Dhruv Batra.
\newblock {E}mbodied {Q}uestion {A}nswering.
\newblock In {\em CVPR}, 2018.

\bibitem{feichtenhofer2016convolutional}
Christoph Feichtenhofer, Axel Pinz, and Andrew Zisserman.
\newblock Convolutional two-stream network fusion for video action recognition.
\newblock In {\em CVPR}, 2016.

\bibitem{fried2018speaker}
Daniel Fried, Ronghang Hu, Volkan Cirik, Anna Rohrbach, Jacob Andreas,
  Louis-Philippe Morency, Taylor Berg-Kirkpatrick, Kate Saenko, Dan Klein, and
  Trevor Darrell.
\newblock Speaker-follower models for vision-and-language navigation.
\newblock In {\em NeurIPS}, 2018.

\bibitem{goodale1992separate}
Melvyn~A Goodale and A~David Milner.
\newblock Separate visual pathways for perception and action.
\newblock In {\em Trends Neurosci.}, 1992.

\bibitem{gordon2018iqa}
Daniel Gordon, Aniruddha Kembhavi, Mohammad Rastegari, Joseph Redmon, Dieter
  Fox, and Ali Farhadi.
\newblock Iqa: Visual question answering in interactive environments.
\newblock In {\em CVPR}, 2018.

\bibitem{he2017mask}
Kaiming He, Georgia Gkioxari, Piotr Doll{\'a}r, and Ross Girshick.
\newblock Mask r-cnn.
\newblock In {\em ICCV}, 2017.

\bibitem{hu2017modeling}
Ronghang Hu, Marcus Rohrbach, Jacob Andreas, Trevor Darrell, and Kate Saenko.
\newblock Modeling relationships in referential expressions with compositional
  modular networks.
\newblock In {\em CVPR}, 2017.

\bibitem{hu2016natural}
Ronghang Hu, Huazhe Xu, Marcus Rohrbach, Jiashi Feng, Kate Saenko, and Trevor
  Darrell.
\newblock Natural language object retrieval.
\newblock In {\em CVPR}, 2016.

\bibitem{jain2021gridtopix}
Unnat Jain, Iou-Jen Liu, Svetlana Lazebnik, Aniruddha Kembhavi, Luca Weihs, and
  Alexander Schwing.
\newblock Gridtopix: Training embodied agents with minimal supervision.
\newblock arXiv:2105.00931, 2021.

\bibitem{Jia2016DynamicFN}
Xu Jia, Bert~De Brabandere, Tinne Tuytelaars, and Luc Gool.
\newblock Dynamic filter networks.
\newblock In {\em NeurIPS}, 2016.

\bibitem{ke2019tactical}
Liyiming Ke, Xiujun Li, Yonatan Bisk, Ari Holtzman, Zhe Gan, Jingjing Liu,
  Jianfeng Gao, Yejin Choi, and Siddhartha Srinivasa.
\newblock Tactical rewind: Self-correction via backtracking in
  vision-and-language navigation.
\newblock In {\em CVPR}, 2019.

\bibitem{ai2thor}
Eric Kolve, Roozbeh Mottaghi, Winson Han, Eli VanderBilt, Luca Weihs, Alvaro
  Herrasti, Daniel Gordon, Yuke Zhu, Abhinav Gupta, and Ali Farhadi.
\newblock {AI2-THOR: An Interactive 3D Environment for Visual AI}.
\newblock arXiv:1712.05474, 2017.

\bibitem{krantz2020navgraph}
Jacob Krantz, Erik Wijmans, Arjun Majumdar, Dhruv Batra, and Stefan Lee.
\newblock Beyond the nav-graph: Vision-and-language navigation in continuous
  environments.
\newblock arXiv:2004.02857, 2020.

\bibitem{landi2019embodied}
Federico Landi, Lorenzo Baraldi, Massimiliano Corsini, and Rita Cucchiara.
\newblock Embodied vision-and-language navigation with dynamic convolutional
  filters.
\newblock In {\em BMVC}, 2019.

\bibitem{Li2019RobustNW}
Xiujun Li, Chunyuan Li, Qiaolin Xia, Yonatan Bisk, Asli Çelikyilmaz, Jianfeng
  Gao, Noah~A. Smith, and Yejin Choi.
\newblock Robust navigation with language pretraining and stochastic sampling.
\newblock In {\em EMNLP/IJCNLP}, 2019.

\bibitem{Liu2018SibNetSC}
Sheng Liu, Zhou Ren, and Junsong Yuan.
\newblock Sibnet: Sibling convolutional encoder for video captioning.
\newblock In {\em ACM MM}, 2018.

\bibitem{ma2019selfmonitoring}
Chih-Yao Ma, Jiasen Lu, Zuxuan Wu, Ghassan AlRegib, Zsolt Kira, Richard Socher,
  and Caiming Xiong.
\newblock Self-monitoring navigation agent via auxiliary progress estimation.
\newblock In {\em ICLR}, 2019.

\bibitem{ma2019regretful}
Chih-Yao Ma, Zuxuan Wu, Ghassan AlRegib, Caiming Xiong, and Zsolt Kira.
\newblock The regretful agent: Heuristic-aided navigation through progress
  estimation.
\newblock In {\em CVPR}, 2019.

\bibitem{MacMahon2006WalkTT}
Matt MacMahon, Brian Stankiewicz, and Benjamin Kuipers.
\newblock Walk the talk: Connecting language, knowledge, and action in route
  instructions.
\newblock In {\em AAAI}, 2006.

\bibitem{habitat19iccv}
{Manolis Savva*}, {Abhishek Kadian*}, {Oleksandr Maksymets*}, Yili Zhao, Erik
  Wijmans, Bhavana Jain, Julian Straub, Jia Liu, Vladlen Koltun, Jitendra
  Malik, Devi Parikh, and Dhruv Batra.
\newblock Habitat: {A} {P}latform for {E}mbodied {AI} {R}esearch.
\newblock In {\em ICCV}, 2019.

\bibitem{misra2017mapping}
Dipendra Misra, John Langford, and Yoav Artzi.
\newblock Mapping instructions and visual observations to actions with
  reinforcement learning.
\newblock In {\em EMNLP}, 2017.

\bibitem{ngyuen_eval_winner}
Van-Quang Nguyen and Takayuki Okatani.
\newblock A hierarchical attention model for action learning from realistic
  environments and directives.
\newblock {\em ECCV EVAL Workshop, https://askforalfred.com/EVAL/}, 2020.

\bibitem{saha2021modular}
Homagni Saha, Fateme Fotouhif, Qisai Liu, and Soumik Sarkar.
\newblock A modular vision language navigation and manipulation framework for
  long horizon compositional tasks in indoor environment.
\newblock arXiv:2101.07891, 2021.

\bibitem{shridhar2020alfred}
Mohit Shridhar, Jesse Thomason, Daniel Gordon, Yonatan Bisk, Winson Han,
  Roozbeh Mottaghi, Luke Zettlemoyer, and Dieter Fox.
\newblock Alfred: A benchmark for interpreting grounded instructions for
  everyday tasks.
\newblock In {\em CVPR}, 2020.

\bibitem{ALFWorld20}
Mohit Shridhar, Xingdi Yuan, Marc-Alexandre C\^ot\'e, Yonatan Bisk, Adam
  Trischler, and Matthew Hausknecht.
\newblock {ALFWorld: Aligning Text and Embodied Environments for Interactive
  Learning}.
\newblock arXiv:2010.03768, 2020.

\bibitem{Simonyan2014TwoStreamCN}
Karen Simonyan and Andrew Zisserman.
\newblock Two-stream convolutional networks for action recognition in videos.
\newblock In {\em NeurIPS}, 2014.

\bibitem{tan2019learning}
Hao Tan, Licheng Yu, and Mohit Bansal.
\newblock Learning to navigate unseen environments: Back translation with
  environmental dropout.
\newblock In {\em NAACL}, 2019.

\bibitem{tesfaldet2018two}
Matthew Tesfaldet, Marcus~A Brubaker, and Konstantinos~G Derpanis.
\newblock Two-stream convolutional networks for dynamic texture synthesis.
\newblock In {\em CVPR}, 2018.

\bibitem{wang2016learning}
Liwei Wang, Yin Li, and Svetlana Lazebnik.
\newblock Learning deep structure-preserving image-text embeddings.
\newblock In {\em CVPR}, 2016.

\bibitem{wang2018look}
Xin Wang, Wenhan Xiong, Hongmin Wang, and William Yang~Wang.
\newblock Look before you leap: Bridging model-free and model-based
  reinforcement learning for planned-ahead vision-and-language navigation.
\newblock In {\em ECCV}, 2018.

\bibitem{zhu2017visual}
Yuke Zhu, Daniel Gordon, Eric Kolve, Dieter Fox, Li Fei-Fei, Abhinav Gupta,
  Roozbeh Mottaghi, and Ali Farhadi.
\newblock Visual semantic planning using deep successor representations.
\newblock In {\em ICCV}, 2017.

\bibitem{Zitnick2014EdgeBL}
Charles~Lawrence Zitnick and Piotr Doll{\'a}r.
\newblock Edge boxes: Locating object proposals from edges.
\newblock In {\em ECCV}, 2014.

\end{thebibliography}
}

\clearpage

{\hspace{-1.5em} \Large \textbf{Appendix}}
\begin{appendix}
\definecolor{Gray}{gray}{0.90}
\newcolumntype{a}{>{\columncolor{Gray}}r}
\newcolumntype{b}{>{\columncolor{Gray}}c}

\newcommand{\bmp}[1]{\textcolor{blue}{#1}} 
\newcommand{\orange}[1]{\textcolor{orange}{#1}} 

\noindent\textbf{Note:} All {\color{orange}orange} characters indicate the index of the main paper. 

\section{ALFRED Benchmark Details} 
We provide the detailed description of the ALFRED benchmark here.
Each expert ground-truth trajectory consists of a set of egocentric visual observation and ground-truth action pairs with corresponding natural language descriptions.
We denote each trajectory by a tuple of $[\{(I_t, a_t)\}_{t=1}^T, S]$ where each $I_t$ and $a_t$ denotes the egocentric observation and the ground-truth action at the time step, $t$.
$T$ is the length of a trajectory and $S$ is the natural language description.
The natural language description, $S$, is composed of a goal statement, $S_{goal}$, and step-by-step instructions, $S_{instr}$. 
The goal statement describes the overall task the agent must complete. The step-by-step instructions provide detailed descriptions on how the agent can accomplish the task.
For more information, please refer to~\orange{[33]}.

Based on the egocentric observations and the language descriptions, the agent predicts an action and a mask for each time step.
The action space is comprised of 5 navigation actions: \textsc{MoveAhead}, \textsc{RotateRight}, \textsc{RotateLeft}, \textsc{RotateRight}, \textsc{LookUp}, and \textsc{LookDown}, and 7 interaction actions: \textsc{Pickup}, \textsc{Put}, \textsc{Open}, \textsc{Close}, \textsc{ToggleOn}, \textsc{ToggleOff}, and \textsc{Slice} along with the \textsc{Stop} action to terminate an episode.
In case of interaction actions, the agent must localise objects of interest. 

\begin{table}[h!]
    \centering
    \resizebox{0.49\textwidth}{!}{
        \begin{tabular}{@{}lacacacac@{}}
            \toprule
            \multirow{2}{*}{Task-Type}
                             & \multicolumn{2}{c}{{Shridhar~\etal~\orange{[33]}}}   & \multicolumn{2}{c}{MOCA}
                             \\
                             & \multicolumn{1}{a}{Seen} & \multicolumn{1}{c}{Unseen} 
                             & \multicolumn{1}{a}{Seen} & \multicolumn{1}{c}{Unseen} 
                             \\
                             
            \midrule
            {Pick \& Place}      & $7.0$ & $0.0$ & $\B{29.6}$ & $\B{6.0}$ \\
            {Cool \& Place}      & $4.0$ & $0.0$ & $\B{32.5}$ & $\B{2.8}$ \\
            {Stack \& Place}     & $0.9$ & $0.0$ & $\B{6.1}$  & $\B{6.4}$ \\
            {Heat \& Place}      & $1.9$ & $0.0$ & $\B{31.8}$ & $\B{5.1}$ \\
            {Clean \& Place}     & $1.8$ & $0.0$ & $\B{30.4}$ & $\B{10.6}$ \\
            {Examine}            & $9.6$ & $0.0$ & $\B{31.9}$ & $\B{4.6}$ \\
            {Pick Two \& Place}  & $0.8$ & $0.0$ & $\B{19.4}$ & $\B{1.2}$ \\
            \midrule
            {Average}           & $3.7$ & $0.0$ & $\B{26.0}$ & $\B{5.2}$ \\
            \bottomrule
        \end{tabular}
    }
    \vspace{-0.5em}
    \caption{
        \textbf{Success rates across 7 task types in ALFRED.}
        All values are in percentage.
        The agent is evaluated on the Validation set.
        Highest values per fold are indicated in \B{blue}.
    }
    \vspace{-0.5em}
    \label{tab:abla_task}
\end{table}

\begin{table}[h!]
    \centering
    \resizebox{0.45\textwidth}{!}{
        \begin{tabular}{@{}lacacacac@{}}
            \toprule
            \multirow{2}{*}{Subgoal}
                            & \multicolumn{2}{c}{{Shridhar~\etal~\orange{[33]}}}   & \multicolumn{2}{c}{MOCA}
                            \\
                            & \multicolumn{1}{a}{Seen} & \multicolumn{1}{c}{Unseen} 
                            & \multicolumn{1}{a}{Seen} & \multicolumn{1}{c}{Unseen} 
                            \\
                             
            \midrule
            {Goto}   & $51$ & $22$ & $\B{54}$ & $\B{32}$ \\
            {Pickup} & $32$ & $21$ & $\B{53}$ & $\B{44}$ \\
            {Put}    & $\B{81}$ & $\B{46}$ & $62$ & $39$ \\
            {Cool}   & $\B{88}$ & $\B{92}$ & $87$ & $38$ \\
            {Heat}   & $\B{85}$ & $\B{89}$ & $84$ & $86$ \\
            {Clean}  & $\B{81}$ & $57$ & $79$ & $\B{71}$ \\
            {Slice}  & $25$ & $12$ & $\B{51}$ & $\B{55}$ \\
            {Toggle} & $\B{100}$ & $\B{32}$ & $93$ & $11$ \\
            \midrule
            {Average}    & $68$ & $46$ & $\B{70}$ & $\B{47}$ \\
            \bottomrule
        \end{tabular}
    }
    \vspace{-0.5em}
    \caption{
        \textbf{Subgoal success rate.}
        The highest values per fold and task are shown in \B{blue}. Note all values correspond to Path-Length-Weighted success rate metric.
    }
    \vspace{-1.3em}
    \label{tab:abla_subgoal}
\end{table}

\begin{table*}[t!]
    \centering
    \begin{small}
        \begin{tabular}{@{}cccccbbcc@{}}
            \toprule
            \multicolumn{1}{c}{\bf~~~~~~~~} & \multicolumn{4}{c}{\bf Components} & \multicolumn{2}{c}{\bf Val-Seen} & \multicolumn{2}{c}{\bf Val-Unseen} \\
            \cmidrule(lr){1-1} \cmidrule(lr){2-5} \cmidrule(lr){6-7} \cmidrule(lr){8-9}
            \multirow{1}{*}{~~~\#} & FPP & OCL & DF & DA & \multirow{1}{*}{Task}   & \multirow{1}{*}{Goal-Cond.} &  \multirow{1}{*}{Task} &  \multirow{1}{*}{Goal-Cond.} \\
            \cmidrule(lr){1-1} \cmidrule(lr){2-5} \cmidrule(lr){6-7} \cmidrule(lr){8-9}
            \multicolumn{1}{c}{($a$)} & \cmark & \cmark & \cmark & \cmark                                               & $25.85$ ($18.95$) & $34.92$ ($26.44$) & $5.36$ ($3.19$)   & $16.18$ ($10.44$)  \\%
            \multicolumn{1}{c}{($b$)} & \multicolumn{1}{c}{} & \cmark & \cmark & \cmark                                 & $22.32$ ($16.17$) & $30.82$ ($23.84$) & $4.51$ ($2.59$)   & $16.65$ ($10.75$)  \\%
            \cmidrule(lr){1-1} \cmidrule(lr){2-5} \cmidrule(lr){6-7} \cmidrule(lr){8-9}
            \multicolumn{1}{c}{($c$)} & \cmark & \cmark & \cmark & \multicolumn{1}{c}{}                                 & $15.85$ ($10.02$) & $23.19$ ($15.78$) & $2.92$ ($1.35$)   & $12.78$ ($6.84$)  \\%
            \multicolumn{1}{c}{($d$)} & \multicolumn{1}{c}{} & \cmark & \cmark & \multicolumn{1}{c}{}                   & $12.56$ ($7.05$)  & $21.29$ ($13.33$) & $2.68$ ($1.32$)   & $13.49$ ($7.63$)  \\%
            \multicolumn{1}{c}{($e$)} & \cmark & \cmark & \multicolumn{1}{c}{} & \multicolumn{1}{c}{}                   & $14.63$ ($9.80$)  & $25.56$ ($18.32$) & \multicolumn{1}{r}{$2.19$ ($1.23$)}   & $10.76$ ($7.36$)  \\%
            \multicolumn{1}{c}{($f$)} & \multicolumn{1}{c}{} & \cmark & \multicolumn{1}{c}{} & \multicolumn{1}{c}{}     & $11.71$ ($5.42$)  & $20.06$ ($11.21$) & $1.83$ ($0.82$)   & $11.04$ ($6.23$)  \\
            \multicolumn{1}{c}{($g$)} & \cmark & \multicolumn{1}{c}{} & \cmark & \multicolumn{1}{c}{}                   & $3.90$ ($2.40$)   & $11.00$ ($7.20$)  & $0.50$ ($0.30$)   & \multicolumn{1}{r}{~~~~~$7.80$ ($4.40$)}   \\%
            \multicolumn{1}{c}{($h$)} & \multicolumn{1}{c}{} & \multicolumn{1}{c}{} & \cmark & \multicolumn{1}{c}{}     & $3.30$ ($1.70$)  & $10.20$ ($6.10$) & $0.40$ ($0.20$)   & \multicolumn{1}{r}{$8.00$ ($4.00$)}\\%
            \bottomrule
        \end{tabular}
        \label{fig:dataset_comparison}
    \end{small}
    \vspace{-0.5em}
    \caption{
        \textbf{Ablation Study for Each Component of MOCA.}
        FPP denotes factorized perception and policy.
        OCL denotes object-centric localisation.
        DF denotes language-guided dynamic filters.
        DA denotes data augmentation.
        For each metric, we report task success rates with corresponding path weighted scores in parentheses.
        The absence of checkmark denotes that the corresponding component is removed.
    }
    \vspace{-1.3em}
    \label{tab:ablation}
\end{table*}

\section{Implementation Details} 
The egocentric visual observations are resized to $224\times224$. 
For the visual encoder, we use a pre-trained ResNet-18~\orange{[18]}.
For the experimental results and analysis in subsequent sections, we use the goal statement as input for the IPM and step-by-step instructions for the APM, otherwise stated (\orange{Sec.~4.2}). 

The model is trained end-to-end using Adam for 30 epochs with an initial learning rate of $10^{-3}$ with a batch size of 16. 
We also use a dropout of 0.2 for visual features and LSTM decoder hidden states. 
We adopt data augmentation for the egocentric observations, $\{I_t\}_{t=1}^T$, to address the sample insufficiency of imitation learning in each trajectory. 
Specifically, we exploit two augmentation methods; color swapping and AutoAugment~\orange{[10]}.

Color swapping randomizes the order of the RGB channels of each frame, which yields 6 combinations in total.
We randomly pick 3 of them, including the original.
AutoAugment randomizes each frame with predefined image operations such as rotation, shearing, and auto-contrast.
We specifically adopt the augmentation policy found for ImageNet. 
For the details of the policy, please refer to \orange{[10]}.

Each augmentation method generates two perturbed trajectories for each trajectory in training our agent.
This results in one original trajectory with four augmented ones (\ie, five training trajectories in total).

\section{Task Type and Subgoal Ablations} 
Tasks in ALFRED~\orange{[33]} are divided into 7 high-level categories.
Table~\ref{tab:abla_task} shows the performance of our factorized agent on each task type.
On short-horizon tasks such as \textbf{Pick \& Place} and \textbf{Examine}, Shridhar \etal~\orange{[33]} which is a single-branch model succeeds in some trajectories in seen environments, but has near zero unseen success rates.
However, our agent outperforms them in both seen and unseen scenes by large margins. 
\textbf{Stack \& Place} and \textbf{Pick Two \& Place} are the two most complex and the long tasks in ALFRED.
Our agent achieves 6.1\% and 19.4\% seen success rates as compared to 0.9\% and 0.8\% of Shridhar \etal.
It also achieves improved success rates in unseen scenes whereas Shridhar \etal show zero unseen success rates.

Following~\orange{[33]}, we also examine the performance of our agent on individual subgoals. 
For the subgoal analysis, we use the expert trajectory to move the agent to the starting time step of the respective subgoal.
Then, the agent starts inference based on the current observations.
Table~\ref{tab:abla_subgoal} shows the agent's performance on individual subgoals.

The \textbf{Goto} subgoal is indicative of the navigation ability of an agent.
Even though navigation in visually complex and unseen environments is more challenging, our model achieves $32\%$ as opposed to $22\%$ of Shridhar \etal.
Although the gap between average subgoal performance of Shridhar \etal and our agent is relatively small, our agent drastically outperforms it on full task completion as shown in \orange{Table 1} of the main paper.
This indicates our agent's ability to succeed on overall task completion and not limiting itself to memorizing short term subgoals only.

\section{Model Component Ablation} 
We provide more results about \orange{Table 3} of the main paper including goal-condition metrics for \textit{Model Ablations} of our agent in Table~\ref{tab:ablation}.
We investigate the significance of each component in detail.
The analysis can be found in \orange{Model Ablations} in \orange{Section 4.2} in the main paper.

\section{Additional Qualitative Examples} 
We present qualitative examples (both successes and failures) of our factorized agent and contrast it with the single-branch model by Shridhar \etal in the accompanied videos.
Each frame in the videos shows the goal statement and step-by-step instructions.
The step-by-step instruction that the agent tries to accomplish at the current time step is highlighted in yellow.
When our agent performs interaction, the predicted target class of the object at that time step is shown on the top-left corner of the egocentric frame.
Note that we do not show object class for Shridhar \etal since they produce class-agnostic masks.

In \textit{success\_1.mp4}, while the method by Shridhar \etal fails to navigate to right object (yellow spray bottles), our agent successfully navigates and places both of them on top of the toilet, thereby satisfying the goal statement. 
It implies that our Action Policy Module (APM) is able to predict accurate action sequences based on vision and language inputs.

For \textit{success\_2.mp4}, both our agent and the prior work navigate correctly to the right locations at various stages of the task.
However, when the instruction asks to pick up the lettuce, our agent correctly localises and picks up the correct object. 
The Interactive Perception Module (IPM) of our agent which enables it to reason about object classes helps it to predict the mask of the correct object (lettuce).
On the contrary, the prior work picks up a cup which was not mentioned in the instruction at all, thereby failing on the tasks even though it performs all the other actions accurately.
This can be attributed to its class-agnostic nature of interaction mask prediction.

Similarly in \textit{success\_3.mp4}, while the method by Shridhar \etal fails to pick up the knife, due to an inaccurately localised mask under limited visibility and picks up the spatula instead, our agent correctly picks up the knife and successfully accomplishes the task.

\textit{success\_4.mp4} demonstrates the ability of our agent to perform the tasks in a more efficient manner.
Even though the prior work successfully navigates to the cup, it takes a lot of unnecessary navigation actions which harm the path-length-weighted score considerably.
In addition, after picking up the cup, it fails to navigate further and ends up being stuck at a desk and therefore fails.
If our agent would have faced a similar scenario, our `Obstruction Evasion' module would have helped the agent to evade it.
On the other hand, our agent navigates to the correct objects of interest (the cup, the refrigerator, and a counter) in a more efficient path.
It also performs accurate interactions and therefore accomplishes the given task.

For the \textit{fail.mp4} video, the prior work tries to interact with an irrelevant object (cloth), instead of the tissue box and fails at completing the task.
Similarly, our agent also tries to interact with the wrong target object (soap bottle) as it fails to navigate to the right position to observe that object, making it invisible.
This navigational failure misleads the IPM to perceive the soap bottle as a tissue box and therefore tries to place an unintended object on top of the toilet and fails at the task.

\end{appendix}

\end{document}